# Night Time Haze and Glow Removal using Deep Dilated Convolutional Network

Shiba Kuanar *Member IEEE*, K.R. Rao *Fellow IEEE, Dwarikanath Mahapatra Member IEEE,*
Monalisa Bilas *Member IEEE*

**ABSTRACT:**
In this paper, we address the single image haze removal problem in a nighttime scene. The night haze removal is a severely ill-posed problem especially due to the presence of various visible light sources with varying colors and non-uniform illumination. These light sources are of different shapes and introduce noticeable glow in night scenes. To address these effects we introduce a deep learning based DeGlow-DeHaze iterative architecture which accounts for varying color illumination and glows. First, our convolution neural network (CNN) based DeGlow model is able to remove the glow effect significantly and on top of it a separate DeHaze network is included to remove the haze effect. For our recurrent network training, the hazy images and the corresponding transmission maps are synthesized from the NYU depth datasets and consequently restored a high-quality haze-free image. The experimental results demonstrate that our hybrid CNN model outperforms other state-of-the-art methods in terms of computation speed and image quality. We also show the effectiveness of our model on a number of real images and compare our results with the existing night haze heuristic models.

## 1. INTRODUCTION:

Haze is an atmospheric phenomenon where mist, fog, dew, dust, and other tiny particles obscure the clarity of atmosphere and reduce the contrast of day and night time images. The above degradation is mainly due to light scattering phenomena in the atmosphere. The aerosols floating on the atmosphere absorb and scatter light on all directions. As a result, multiple reflected light rays scatter out to all directions other than the line of sight and attenuate the screen reflection with distance. Hence the light rays from light sources scatter into the line of sight of the camera sensor, create airlight (Fig. 1) or veiling light and wash out scene visibility. These combined scattering effects adversely affect the scene visibility which in turn also impacts the subsequent low-level computer vision processing and computational photography applications.

In recent years, numerous daytime haze removal models have been proposed to address the hazy image visibility enhancement. The key to their success mostly relies on the correct estimation of various image priors, and atmospheric light. The standard haze model described by Middleton in [1], [2] illustrates the hazing process as a linear combination of airlight [3] and direct transmission. The direct transmission represents the scene reflection whose intensity reduces by the scattering out process. On the other hand, the airlight represents intensity resulted from the scattering in process of the light sources present on the surrounding atmosphere. Hence the transmission light conveys a fraction of the scene reflection and reaches the camera. Again most of the haze models assume that the atmospheric light present in the input image can be estimated by the brightest image region with a strong approximation. After estimating the atmospheric light, the daytime haze methods calculate the transmission light by using various cues such as dark channel[4], local contrast [5], image fusion [6], and statistical independence between the albedo and shading [7]. The main implementation differences between these methods are due to various cues incorporated with transmission light estimations.

But the effectiveness of these daytime haze methods is not well demonstrated to correct the night time scenes. The main reason might be that the daytime haze model priors do not hold well for most nighttime scenes. The night scenes commonly have numerous and diverse colored light sources, e.g. building, vehicle, street lights etc. which results in non-uniform illumination. These illuminations not only make ambient light estimation inaccurate but also cause some image priors to become invalid. Besides that, night sources introduce more brightness to existing atmospheric light, boost intensity unrealistically and cause the prominent glow to the scene (Fig. 1). The atmospheric light on night scenes are not assumed to be globally uniform and cannot be calculated from the brightest region of night image. As a result, the atmospheric light approximation can differ significantly from that of the brightest intensity in a scene. Consequently, if we normalize the input image w.r.t brightest region intensity then it would cause a prominent color shift in the input image. Our main contributions are summarized as follows.

**Contribution:** To address the nighttime dehazing problem, we introduce a CNN based new haze model that includes the glow effect in addition to the standard haze model described in [3]. As a result, our nighttime atmospheric scattering model can be formally written as a combination of three terms: direct transmission, airlight, and glow. Our deep dilated CNN model learns the image features in two steps and removes the glow effect recursively from night scenes. Our proposed contextual dilated network enlarges the receptive field and retrieves more contextual information on each step (Fig. 3). As a result, the extracted features are enhanced progressively in each recurrence by aggregating information from several parallel dilated convolutions. Our proposed algorithm first decomposes the glow effect from an input image and produces a new hazy image with reduced glow (DeGlow). Then, a separate DeHaze network is included to learn the scene transmission map from the DeGlow image. Our Dehaze network is based on the same contextual dilated network with one recurrence only (Fig. 3). During test time, a spatially varying atmospheric light map is separately estimated from the above transmission map and finally outputs clear scene reflection (Fig. 2 block diagram). To the best of our knowledge, this work is the first that combines deep learning on night dehazing process, for the purpose of high-quality image retrieval. Our estimated scene reflection shows that it has better visibility with reduced haze and glow effect and does not suffer from color shifts.

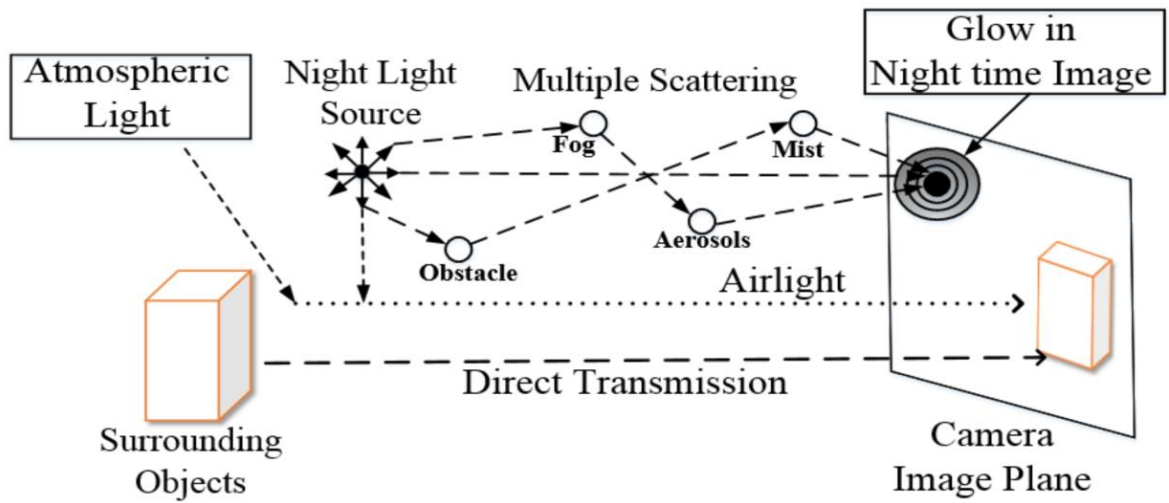

Fig. 1: Block-diagram of our proposed Night-Time Haze base model. Airlight: ambient light reflected from atmospheric objects and aligned into a line of sight. The dark circular structure on image represents the Glow effect of light sources.

In the next section, we review related heuristic and learning based recent approaches on image dehazing. The following sections explain our night haze model design and various evaluation tradeoffs, introduce our architecture with a dilated network for glow/haze removal, and describe our experimental framework. Finally, we demonstrate experimental results on a randomly collected real and synthetic images.

## 2. RELATED WORK:

A variety of approaches has been employed to solve the ill-posed hazing effect from a single daytime image [8], [4], [9], [7], [5], [10], [11]. These heuristic methods use the standard haze model described in [3] and estimate the atmospheric light from the brightest region of an input image. He et al. [4] proposed the famous dark channel prior to estimating the depth map, which achieved good results. Tan et al. [5] produced a haze-free image by maximizing the contrast of the image, but that approach suffered from color distortions. Zhu et al. [12] created a linear model for estimating the scene depth of the hazy image under a color attenuation prior and used supervised learning to estimate model parameters. With the depth map of haze images they estimated the transmission map and restored the scene radiance, and thus effectively removed the haze from a single image. Meng et al. [10] proposed a practical regularization dehazing method to restore the haze-free images by exploring the boundary constraints on the transmission function. Combined with a weighted L1 norm, the optimization problem had a closed form solution in each iteration by using a variable splitting trick. As a result, the proposed algorithm was able to restore a high-quality haze-free image with better colors and fine image details. Based on the color lines pixel regularity in natural images, Fattal et al. [7] presented a single image dehazing method. He derived a local Gaussian Markov random field model that considers the hazy scene regularity and estimate the scene transmission. Kim et al. [13] introduced a dehaze algorithm based on optimized contrast enhancement and proposed a hierarchical quad-tree subdivision search to determine the transmission values. The block-wise contrast was maximized by minimizing information loss due to pixel truncation. Later they extended the algorithm to videos by adding the temporal coherence measures.

More recently, several machine learning techniques, such as random forest regression, decision tree, and convolutional neural networks have been proposed on the domain of image dehazing. Tang et al. [8] introduced a random forest model and investigated different haze relevant features to identify the best feature combinations for image dehazing. Cai et al. [14] adopted a CNN based DehazeNet architecture, which improved the quality of the recovered haze-free image. Most recently, Ren et al. [15] proposed a multi-layer deep neural network to learn the mapping between hazy images. Though these CNN based learning methods achieve superior performance over the recent state-of-the-art methods, they limit their capabilities by learning a mapping between the input hazy image and transmission maps. This is mainly due to the fact that the haze models assume a constant atmospheric light.

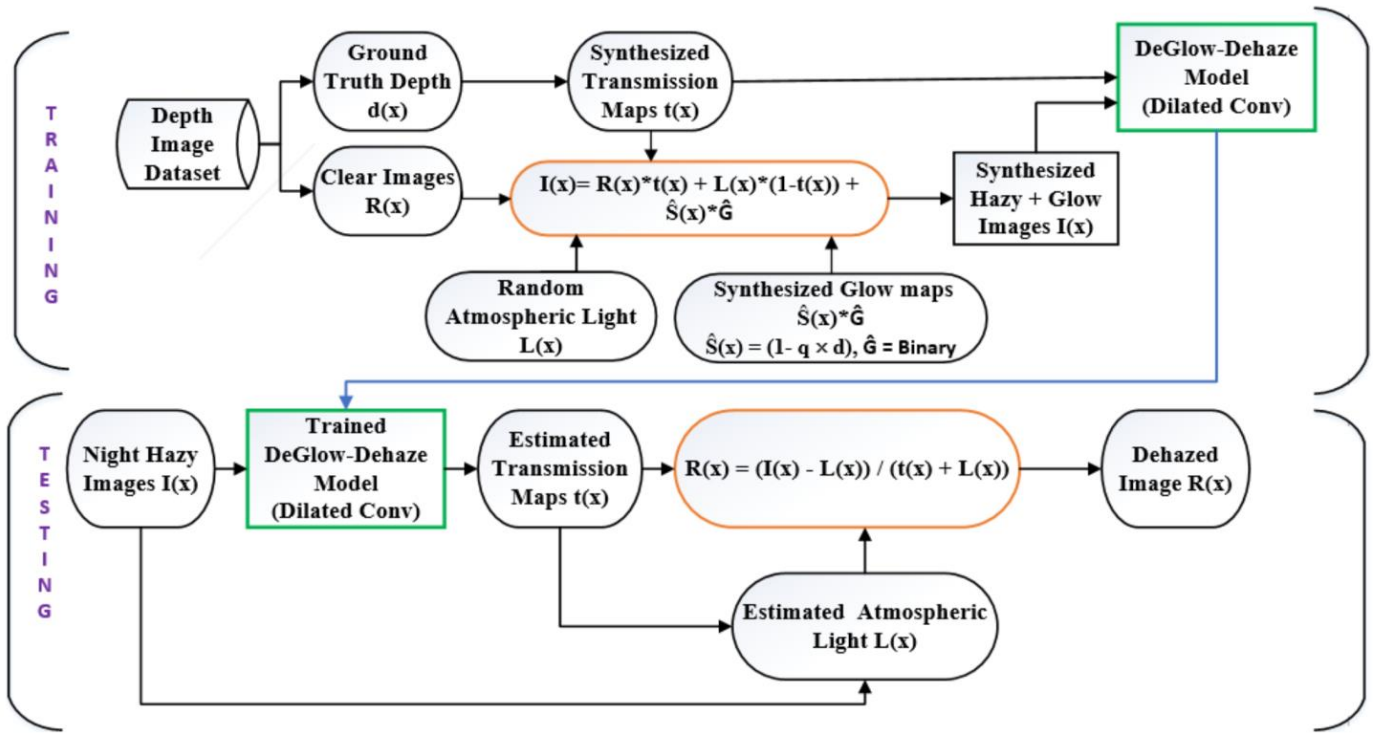

Fig. 2: Synthesized data, Training and Testing steps of our proposed nighttime single image "DeGlow-DeHaze" model.

However, these methods may not be applicable for night hazy images due to non-uniform and multicolored illumination during nighttime. There has been a relatively small number of published methods about night haze removal in the past decade. Pei and Lee [16] proposed a color transfer method on a separate preprocessed step and mapped colors of a night haze image on to a daytime haze image using a dark channel prior technique. Zhang et al. [17] proposed a nighttime dehazing method by including color correction, histogram stretching and gamma correction post-processing steps for final output image enhancement. Srinivasa et al. [18] analyzed the glow effect of light sources and modeled the glow effect as an atmospheric point spread function (APSF). Li et al. [19] modified the daytime haze model by adding an APSF component to the night glow effect and applied a layer separation algorithm to decompose glow from the input image. Recently Ancuti et al. [6] introduced a fusion based approach to enhance the night hazy scene visibility. To the best of our knowledge, no one has implemented a CNN based model on nighttime haze removal. In this paper, a deep dilated CNN based architecture is implemented to jointly detect and remove glow from the night scenes, so as to produce a haze free image with better visibility.

### 3. NIGHTTIME HAZE MODEL:

Fig. 1, displays the standard night haze model diagram with the presence of night light sources. In our model, the atmospheric airlight is assumed to be globally uniform and contributes to the brightness of the light. The direct transmission term describes the light traveling from object reflection and making its way to camera image plane. In addition to airlight and direct transmission, the model also includes a glow term. The glow effect represents the light from night sources which gets scattered multiple times from different directions and reaches the camera plane. These night light sources have varying colors and illumination and potentially contribute to the appearance of airlight. In the nighttime haze environment, the radiated lights from light sources scatter in all arbitrary directions and change smoothly in space. These scattered lights aggregate and lead to smoothly changing light. As a result, the active light sources generate glow on nighttime scenes with the presence of substantial atmospheric particles. We reviewed the recent night dehaze models [19], [16], [6], [18] and generalize to include glow term in our night haze CNN model formulation. The model incorporates two terms 1) glow detection and 2) glow removal. We formulate the haze model equation as below:

$$I(x) = \underbrace{\overbrace{R(x) * t(x)}^{\substack{\text{Direct Transmission} \\ \text{Attenuation}}} + \overbrace{L(x) * (1 - t(x))}^{\text{Airlight}}}_{\text{Glow Contaminated Haze Image}} + \overbrace{\sum_{k=1}^{n} \widehat{S}_k(x) * \widehat{G}}^{\text{Glow Term}} \quad (1)$$

Formulation for classic daytime haze model

In the above equation, I(x) is observed haze image at each pixel position x, and R(x) is clear scene reflection in the absence of glow or haze particles. The transmission map $t(x) = e^{-\beta \cdot d(x)}$ indicates the portion of reflection that has not scattered out and reaches the camera sensor. Term $\beta$ is the scattering coefficient of the atmosphere, and represents the scene depth between the camera and the object. The term L(x) is the global atmospheric light and $\{L(x) \cdot (1-t(x))\}$ represents the particle veil induced by scattering in the process. We incorporate $\widehat{S}_k(x)$ and $\widehat{G}$ terms in our (1) for our glow analysis. Each $\widehat{S}_k(x)$ term represents the shape and illumination direction of a glow source. The subscript k represents the overlapping glow numbers ranging from1 ton, where n is the number of atmospheric glow sources. The variable $\widehat{G}$ indicates the region of visible glow spots on the image and takes a binary mask where 1 indicates Glow region and 0 non Glow region. Separately modeling the glow terms, $\widehat{G}$ and $\widehat{S}_k(x)$ provides 2 benefits: i) it facilitates using additional information to detect glow regions, and ii) allows the network to act separately on Glow and non-Glow regions without affecting the background details of haze images.

#### 3.1. Equations for Glow Detection and Removal

In the nighttime, the airlight obtains its energy from natural atmospheric lights and also from the active light sources. The strong light from light sources directly travels to the image, and some of that light manages to reach the camera sensor after a collision with the surrounding objects [19]. Because of the above processes, light sources create their own presence in the scene and manifest themselves as glow. Hence the glow detection and removal becomes an essential part of night dehaze applications.

$$I(x) = J(x) + \sum_{k=1}^{n} \widehat{S}_k(x) * \widehat{G} \quad (2)$$

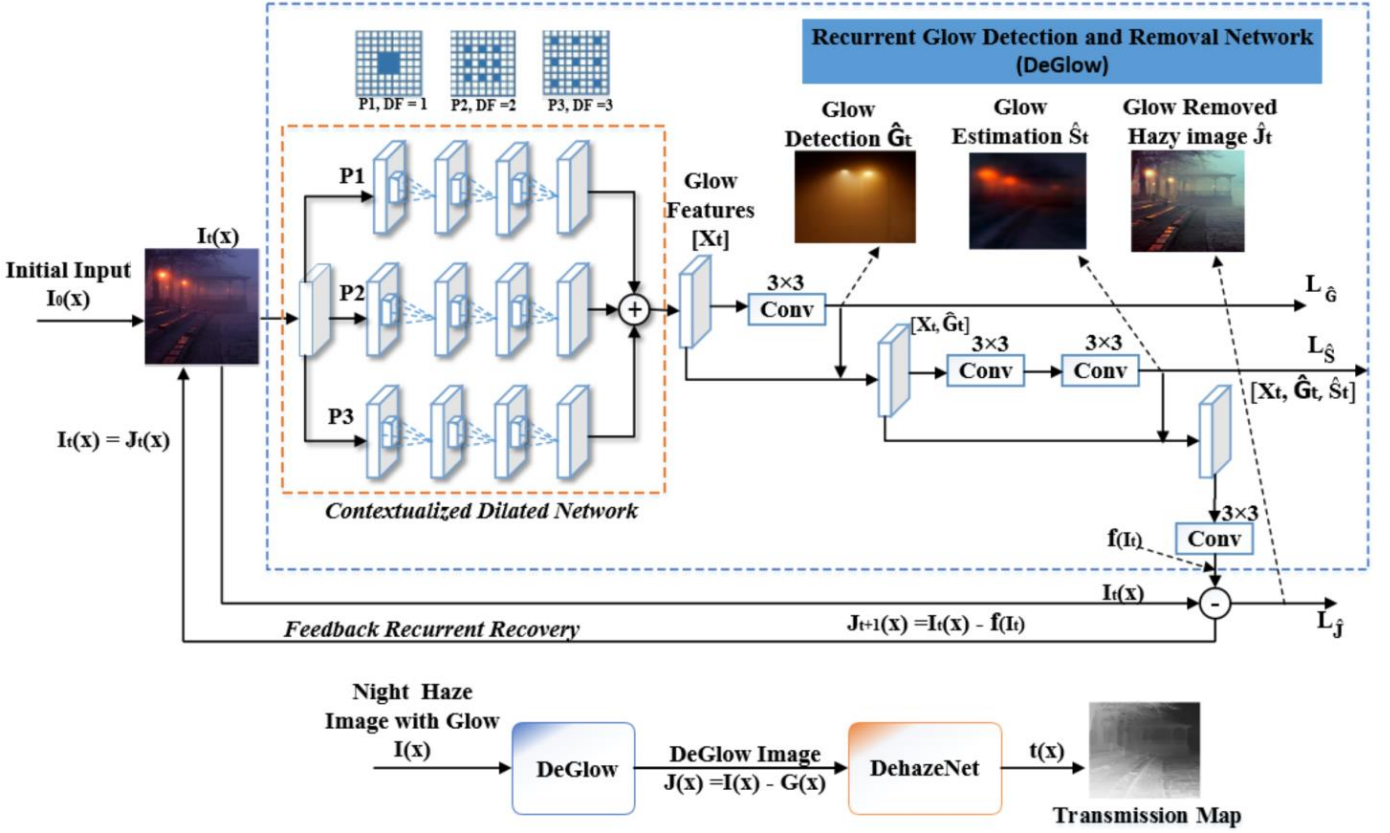

In the above equation, $J(x) = R(x) * t(x) + L(x) * (1−t(x))$ is the night haze image without glow and $\sum_{k=1}^{n} \widehat{S_k}(x) * \widehat{G}$ is the newly included glow term. Our goal is to estimate the scene reflection $R(x)$ for every input image pixel $I(x)$ and recover the haze-free image. First, we detect the glow effect and subsequently remove the glow term from the input image by using our proposed recurrent dilated network (Fig. 3). Later we de-compose the glow image $G(x)$ from $I(x)$ and obtain the hazy image i.e. $J(x) = I(x) – G(x)$. Finally, a standard DeHazed model [14] is used to remove the haze effect from $J(x)$ and obtain the balanced color image. Our end-to-end DeGlow-DeHaze framework (Fig.3) calculates the transmission map $t(x)$. To recover the dehazed image $R(x)$ (Testing phase in Fig. 2), we separately estimate the atmospheric light $L(x)$ from $t(x)$. We follow the steps described in [14], [4] and calculate the atmosphere light $L(x)$ by picking the top 0.1% darkest pixels in a transmission map $t(x)$ [4]. Among the above pixels, one with the highest intensity in $J(x)$ is selected as atmospheric light as shown in Fig. 2.

### 3.2. Contextual Dilated Network

For learning the glow features in our model, the dilated convolution network aggregates contextual information at multiple scales [20]. The motivation of using the dilated convolutions is to support the exponentially expanding receptive fields without losing resolution and to leverage more contextual information without losing local details. Our network gains the contextual information in two steps: I) a recurrent structure which provides an exponentially expanding receptive field on subsequent layers, and II) in each recurrence the output features $X_t$ are obtained by aggregating the representations from three dilated convolution paths. Our model first transforms input night hazy images into a feature space through a series of convolutions (Fig.3). The dilated convolution weighs pixels with a step size equal to the dilated factor (DF), and it increases the receptive field without losing resolution. Three dilated paths P1, P2, and P3 are shown on the Fig. 3 and each consist of three convolutions with a kernel size of 3×3. The above paths uses different DF's i.e. DF =1, DF = 2, DF = 3 and obtain their expanded receptive field as 7×7, 13×13, and 17×17 respectively [20].

### 3.3. CNN Based Recurrent DeGlow Network

A convolutional multi-task network is proposed to remove the glow effect in our model. Based on (2), our network solves an inverse problem through recursive learning. Our dilated convolution extracts the discriminative features, facilitates glow removal from the haze image, and finally calculates the transmission map t(x) as shown in Fig. 3. Given the observed night glow image I(x), our goal is to estimate the J(x), S(x), G(x) terms, and this estimation is an ill-posed problem. To make the problem well-posed and tractable, a joint probability of J(x), S(x), and G(x) is proposed by using the maximum-a-posteriori (MAP) estimation technique. So our MAP output maximizes the probability (J, S, G|I) ∝ P(I|J, S, G) · P(J) · P(S) · P(G), under the assumption that J, S, and G are independent as stated in [21], and [22]. With the algebraic manipulation in the negative log, we follow the below energy minimization equation:

$$arg \min_{J,S,R} ||I - J - \hat{S} - \hat{G}||_2^2 + P_j(J) + P_s(\hat{S}) + P_g(\hat{G}) \qquad (3)$$

Here, $P_j(J)$, $P_s(\hat{S})$, $P_g(\hat{G})$ are enforced layer priors on J, $\hat{S}_k(x)$ , and $\hat{G}$ respectively. The priors are implicitly included to regularize the network and learn from the training datasets. The estimations of J, $\hat{S}_k(x)$, and $\hat{G}$ are intrinsically correlated. Thus, the estimation of I(x) benefits from the recurrent $\hat{S}$, and $\hat{G}$ predictions. We first utilize a dilated network to extract the Glow features $X_t$ from the input images as shown in Fig. 3. The intermediate components are computed as follows: 1) $\hat{L}_G$ is estimated by a 3×3 conv on $X_t$ vector, ii) $\hat{L}_S$ is predicted from two 3×3 convolutions on [X, $\hat{G}$] concatenation vector, and iii) finally $\hat{L}_J$ is computed from a 3×3 conv on [X, $\hat{G}$, $\hat{S}_k(x)$, I $- \hat{G} \times \hat{S}_k(x)$] vector. Above three steps implies a continuous recursive glow removal process. In each iteration, the results from the three paths aggregate along with input features from the previous recurrence through a feedback path.

Our recurrent model can be seen as a cascade of convolutional glow detection networks which performs progressive glow removal and recovers the image with increasingly better visibility. The blue dash box in Fig. 3 exhibits the recursive process of our network. On each recurrence, the residual image between I(x), and J(x) is generated and designated as f (). Subsequently, the glow detection and removal works as follows:

$$[\epsilon_t, G_t, S_t] = f(I_t)$$
$$\epsilon_t = I_t - J_t \ or \ J_t \ = I_t - \epsilon_t \qquad (4)$$
$$I_t + 1 = J_t$$

In (5), the term τ denotes the total number of iterations. In each iteration, the predicted residue f ($I_t$) accumulates and propagates to the final estimation by updating the $I_t$ and $B_t$ values. One can notice that the estimated glow mask $G_t$ and glow steak $S_t$ are not cast directly into the next recurrence. The final estimation can be expressed as:

$$J_\tau = I_0 + \sum_{k=1}^{\tau} \epsilon_t \qquad (5)$$

Hence, the above process removes glow spots progressively based on the intermediate results from the previous steps. The complexity of glow removal is reduced progressively in each iteration by enabling better estimation typically in the presence of heavy night light.

### 3.4. DeGlow Network Training

In our derivation, we introduce three inverse recovery functions as $F_G (\cdot)$, $F_S (\cdot)$, and $F_J (\cdot)$. These functions are modeled by our network, and they respectively generate output $\hat{G}_k$, $\hat{S}_k$, J(x) from the input I(x). We use θ to represent the network weight and bias parameters i.e. θ = {W, b} and n sets of data ($I_k$, $J_k$, $\hat{G}_k$, $\hat{S}_k$) are included for our training, where set k varies from 1 to n. The loss function is parametrized by the θ term so that it jointly estimate G, S, and J based on glow image I.

$$L(\theta) = \frac{1}{n} \sum_{k=1}^{n} ( \overbrace{\|I_k - J_k\|_2^2}^{\text{Direct Error Term}} + \lambda_1 \overbrace{(\|F_S(I_k, \theta) - \hat{S}_k\|^2) + \|F_J(I_k, \theta) - J_k\|^2)}^{\text{Loss Term Due to Priors}}$$

$$- \lambda_2 \overbrace{(log\hat{G}_{k,1} * \hat{G}_k + log(1 - \hat{G}_{k,2})(1 - G_{k,2}))}^{\text{Loss Term Due to Glow (Binary)}}$$

$$where, \; \hat{G}_{k,m} = \underbrace{\frac{\exp\{F_{S,m}(I_k, \theta)\}}{\sum_{k=1}^{2} \exp\{F_{S,i}(I_k, \theta)\}}}_{\text{Glow Region Map Expression}}, \; m \in \{1, 2\}, \; binary$$

$$(6)$$

As the network works recursively, it introduces an additional time variable t to the above loss function $L(\theta)$ in (6). As a result, our time-varying Loss function is represented as $L(\theta_t, t)$. At time $t = 0$, $L(\theta_0, 0)$ is equal to $L(\theta_0)$ and at $t > 1$, $L(\theta_t, t)$ is equivalent to $L(\theta)$ which replaces $I_i$ and $\theta$ by $I_{i,t}$ and $\theta_t$, respectively. The $I_i$, t is generated from the $t_{th}$ iterations of the process given by (4) with an initial value $I_i$. Then, the total loss function $L_{Iter}$ for training $f(I_t)$ is:

$$L_{Iter}(\theta_0, ...\theta_\tau) = \sum_{k=1}^{\tau} L(\theta_t, t) \qquad (7)$$

The cost function in (6) is non-convex due to the presence of priors. We optimize our scheme based on the regularization explained on [23], [24], and [21]. Here $\lambda_1$ and $\lambda_2$ are the weight factors. Loss function $L(\theta)$ is minimized using stochastic gradient descent with standard back-propagation. The weight matrices of each layer are initialized by randomly drawing from a Gaussian distribution with mean = 0 and standard deviation = 0.0001. The bias terms are set to 0. The weight matrices are updated in terms of learning rate and weight gradients. The momentum parameter is set to 0.9. The batch sizes are set to 128 images, and training is regulated by weight decay of 0.001. The learning rate is decreased by a factor of 10 when the validation set accuracy stops improving. Our implementation is derived from the publicly available Matlab/python based Tensor-flow framework [25]. The training is performed on an NVIDIAK40 GPU machine. The Multi-core training operation is carried out by splitting the training images into batches and parallelly processing the task on each core. The network parameters converge after around 1200 iterations. Our entire training process took approximately eighteen hours on our GPU system.

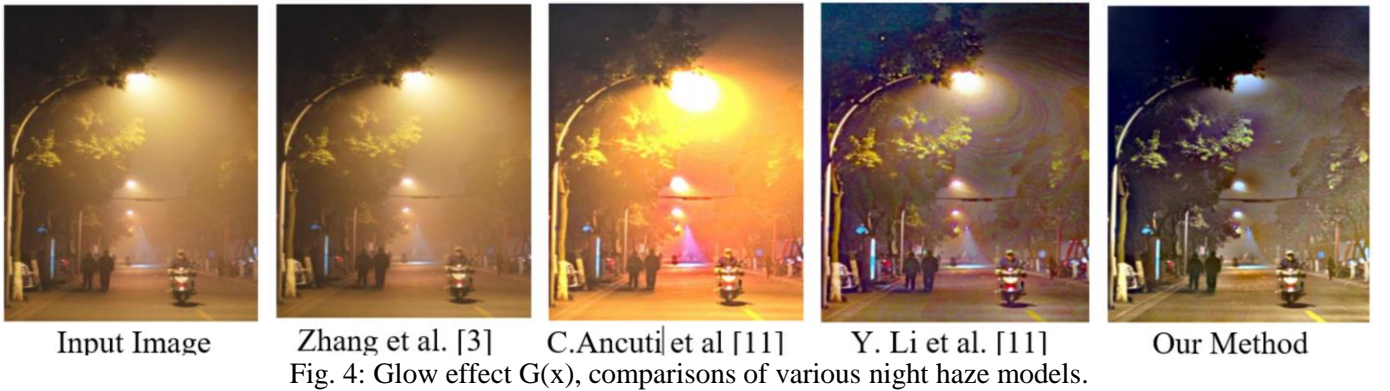

Input Image     Zhang et al. [3]     C.Ancuti et al [11]     Y. Li et al. [11]     Our Method

Fig. 4: Glow effect G(x), comparisons of various night haze models.

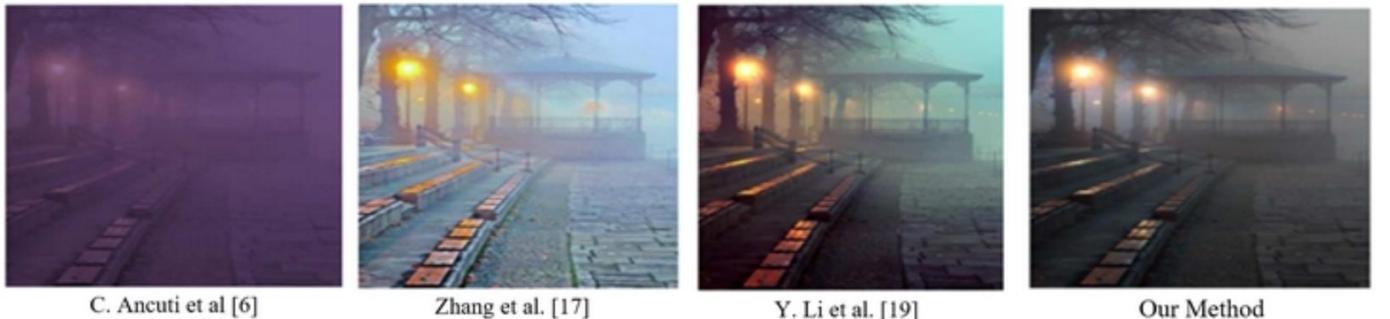

C. Ancuti et al [6]     Zhang et al. [17]     Y. Li et al. [19]     Our Method

Fig. 5: Estimation results of hazy image J(x) from left to right Ancuti et al. [6], Zhang et al. [17], Y. Li et al. [19], and Our Model.

## 3.5. Dehazing with Multi-Scale CNN Network

After the glow effect is removed by our DeGlow model, the haziness still persists on its output J(x). To overcome these haze effects, a separate DeHaze model is created. Our model is based on the structure of the contextualized dilated framework, with one recurrence only. For our network training, we follow the steps discussed in the existing daytime haze model designed by Cai et al. [14]. The training is performed with the same synthesized images (Fig. 3) and estimated scene transmission map t(x). As a result, DeHaze post-processing step suppresses the bright glow effects and enhances the image contrast and visibility.

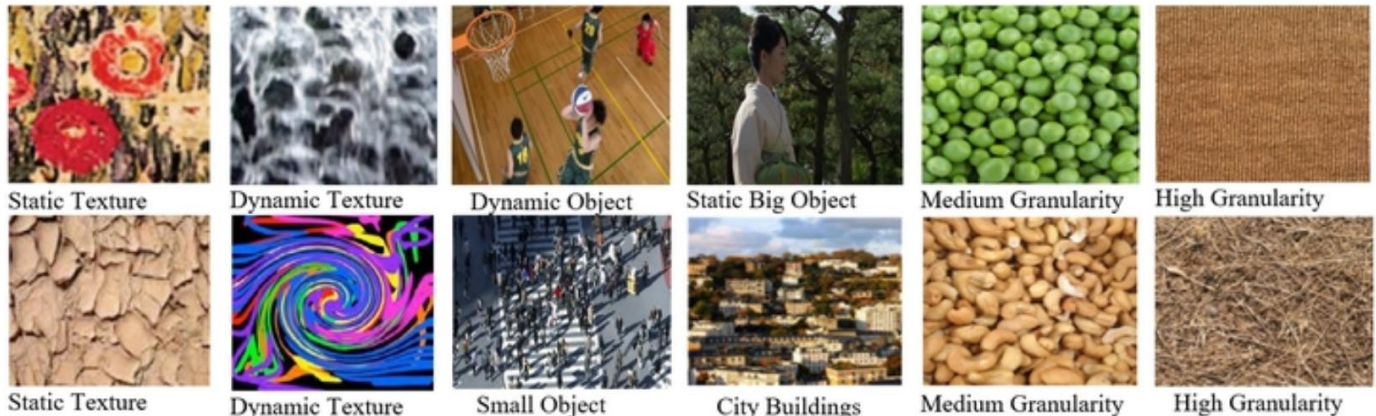

Fig. 6: Example of haze free natural images collected from the Internet

## 4. EXPERIMENT RESULTS:

In general, it is computationally expensive to collect a vast amount of labeled datasets for the deep CNN model training. There are no publicly available large datasets with pairs of hazy image and their associated medium transmission map. Consequently, we resort to synthesizing the training data based on the physical haze formation model analyzed in [15]. Besides that, we also gathered foggy and hazy night images from the publicly available sources with various quality and file formats. For our model parameter learning, we collected a diversified dataset which contains hundreds of images with different textures, object shapes and colors. These hazy images not only capture people's daily life activities but also include those of natural and city landscapes. We believe that this variety of training sample can be learned into the filters of a dilated recurrent network. Fig.6 shows the examples of our collected haze-free images.

Table 1: PSNR and SSIM results on the natural hazy images

| Average Metrics | Haze Image | Pie and Lee's [16] | Zhang et al. [17] | Y. Li et al. [19] | Our Model |
|---|---|---|---|---|---|
| PSNR | 14.11 | 14.98 | 16.13 | 17.66 | 19.11 |
| SSIM | 0.992 | 0.9927 | 0.9958 | 0.9967 | 0.9984 |

We evaluated our proposed CNN algorithm on both synthetic and natural night hazy images and compared our results with various night haze methods. To train our DeGlow-DeHaze model we generated synthesized hazy image I(x) dataset and their transmission maps t(x) (Fig. 3). For training purpose, we randomly sampled 5000 clean images J(x) and corresponding depth maps (d) from NYU Depth dataset [27]. Given the clear image R(x) and ground truth depth, we synthesized the haze image using physical model given on (1). For our validation, we synthesized a set of 2k hazy images from Middlebury stereo database [28], [29], [30]. We generated the random atmospheric light L (t) = [t, t, t], where t ∈ [0.5, 1.0]. Then we sampled three random scattering medium coefficients β ∈ [0.5, 1.5] for every image. We followed the radiative transfer equation on [18] and approximated the attenuation of night glow illumination using a negative exponential form as $\hat{S} = \exp^{-q \times d}$ where variable q was the forward scattering parameter and constant d represented the normalized distance between light source and scene point . Since d was large, we considered the first order Taylor series expansion and approximated as $(1 - d \times q)$. In our experiment, we took the values from q ∈ [0.2, 0.9]. Overall we had 45000 hazy images and transmission maps for training (5000× 3 β terms × 3 q values) and resized images to 320×240 resolution.

Table 2: Average runtime in seconds per image, on various night-haze models.

| Image Resolution | Pie and Lee's[16] | Ancuti et al.[6] | Zhang et al. [17] | Y. Li et al. [19] | Our |
|---|---|---|---|---|---|
| $427 \times 370$ | 10.63 | 7.42 | 5.24 | 6.31 | 0.86 |
| $640 \times 480$ | 19.84 | 14.55 | 8.93 | 9.55 | 1.75 |

The performance of our model was evaluated by using the PSNR and SSIM metrics on test images (Table 1). We collected forty natural images from the newly collected dataset for the average runtime evaluation. The comparative experimental results of various night haze models are shown in Table 2. Again Fig. 4, and 5 show the output glow images G(x) and haze image J(x) respectively. The dehazed results from our model indicate a better estimation of the transmission.

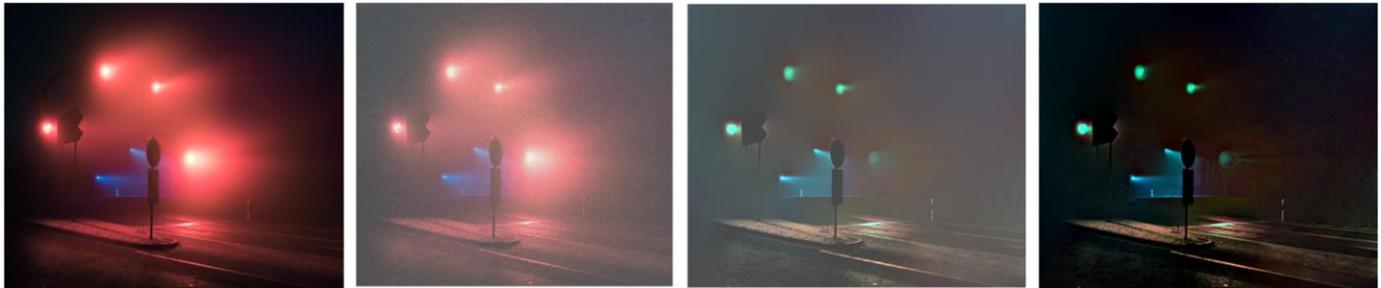

Input Haze and Glow Image    W. Ren et al.[2], (0.874)    B. Cai et al.[6] (0.9831)    Our Method (0.9984)

Fig. 7: Quantitative evaluation by using various Dehaze CNN models on the synthetic image. The large SSIM [26] value of our model implies it's closeness to the ground truth than others.

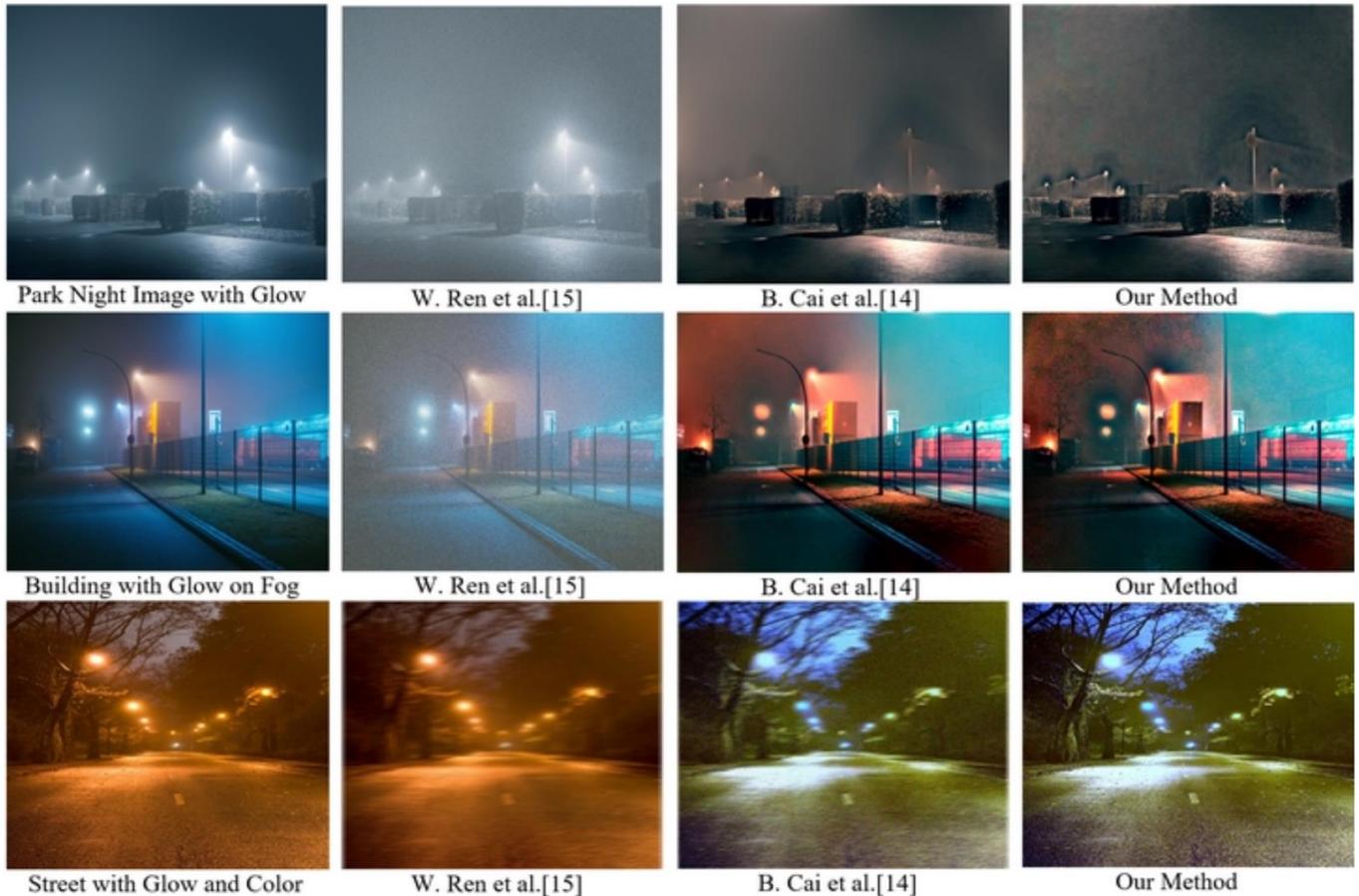

Park Night Image with Glow    W. Ren et al.[15]    B. Cai et al.[14]    Our Method

Building with Glow on Fog    W. Ren et al.[15]    B. Cai et al.[14]    Our Method

Street with Glow and Color    W. Ren et al.[15]    B. Cai et al.[14]    Our Method

Fig. 8: Visual qualitative comparison of our "DeGlow-DeHaze" model w.r.t other DeHaze CNN models. Left and rightmost columns show the input real image and output respectively. The presence of color and glow effect in haze images are reduced essentially and localized.

### 4.1. Quantitative Comparison on Real Night Images

To verify our DeGlow-DeHaze model performance, we presented our results on both real and synthetic night images. The visually inspected results are shown row-wise in Fig. 9, which correspond to Zhang et al. [17], Li et al. [6] and our method respectively. The Zhang et al.'s method generated high intensity in some regions due to an additional post-processing step and introduced the blurring. Though Li et al.'s method was able to remove the night glow effect to a great extent but could not handle color distortion very well around the edges (Fig. 9). It generated fringing artifacts in the surrounding areas of light sources. On the other hand, our proposed approach was able to correct the glow effect, normalized the color distortion and enhanced the visibility with subtle artifacts. Fig. 7 shows the dehazed synthesized images by our approach without much glow artifacts around light sources and with better color balance. Again we performed the quantitative evaluation using SSIM [26] on synthetic images and compared with other CNN models. Our method shows the highest SSIM index, which implies that it is more close to the input ground truth. Besides that, in Fig. 8 we have shown the glow decomposition results with input real images on Left most column. The middle and right columns are showing the estimated haze images by various CNN models i.e. Ren et al. [15], Cai et al. [14] respectively and our method respectively. The presence of glow in haze images are reduced gradually along the row. Overall our proposed algorithm out-performs other CNN models on both real and synthetic image cases.

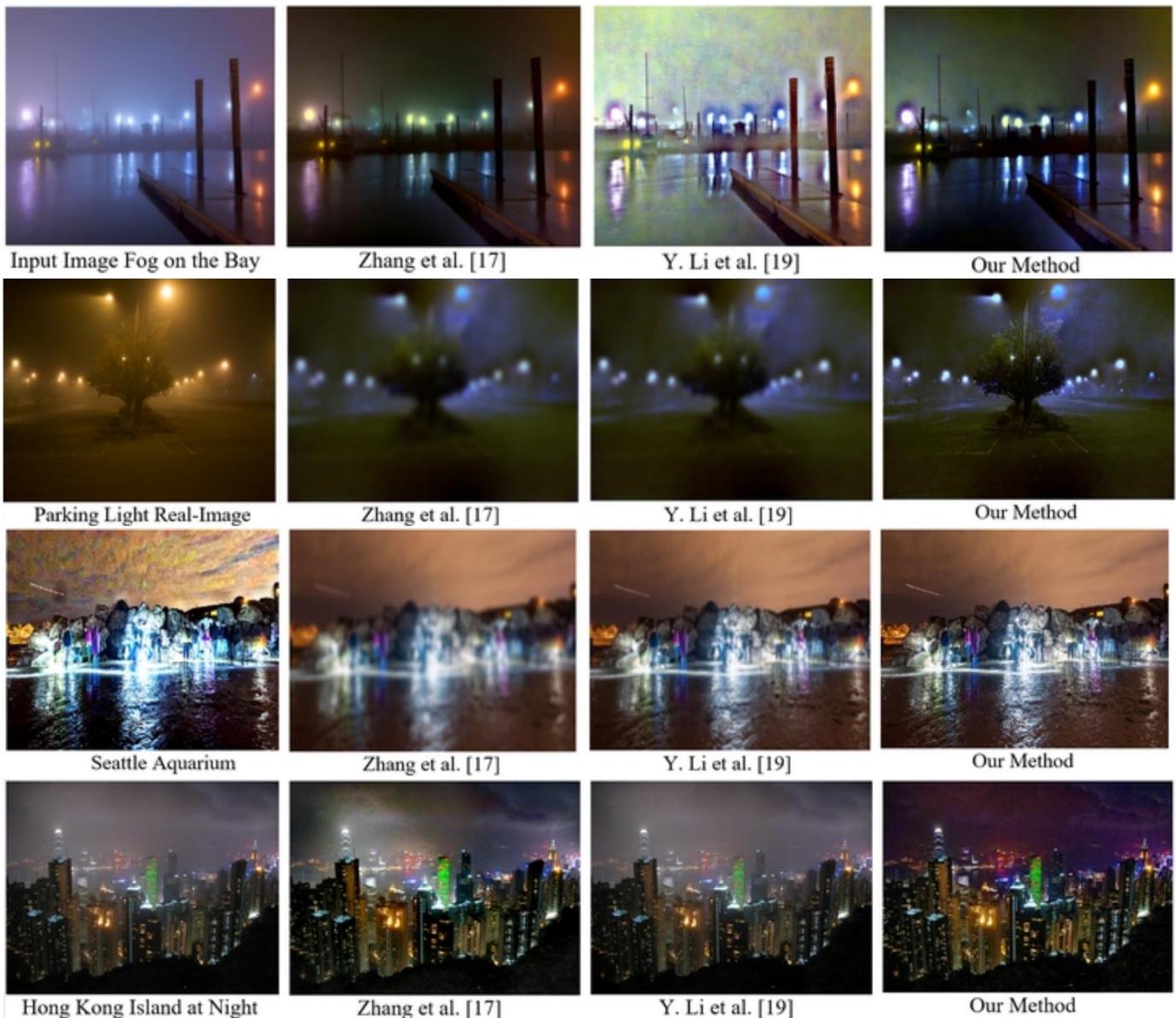

Fig. 9: Visual Comparison of our "DeGlow-DeHaze" results for the Input Real Images.

## 4.2. Failure Case

One failure example is shown in Fig. 10, where our proposed scheme generates color distortions for objects such as the roads and leaves. Another limitation of our algorithm might occur if the objects are surrounded by a large object or the object is occluded by some scene parts. In such situations, our method fails and is shown in Fig. 10. In night haze environment the light radiated from sources changes smoothly in space except at some occlusion positions. The occlusions result in sudden changes between shade areas and bright regions. As a result, these boundaries are very sparse in the whole image. In Fig. 10 d) one can observe that the light glows are merged (behind the bus) and blow out (street light) rather than eliminated. In future work, we would like to address this problem by more advanced color constancy techniques [31], [32] and datasets, and other occluded techniques.

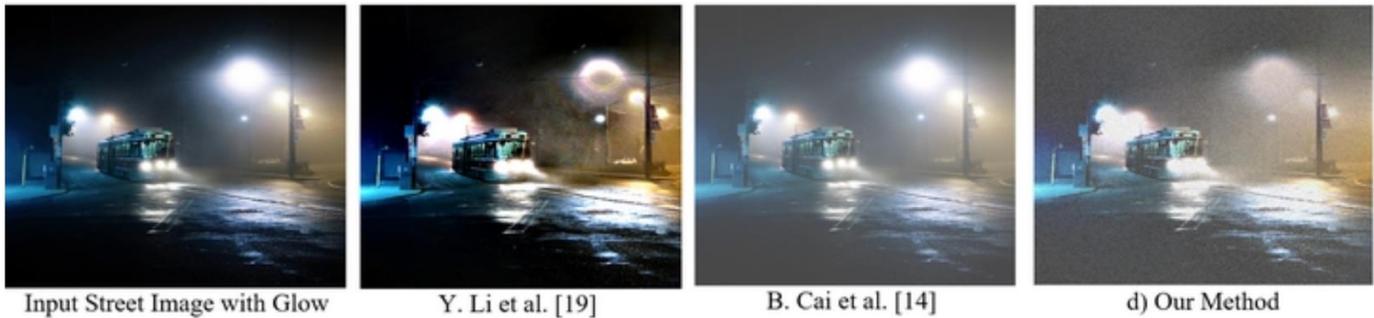

Input Street Image with Glow      Y. Li et al. [19]      B. Cai et al. [14]      d) Our Method

Fig. 10: Failure case for nighttime hazy image on Occlusion.

## 5. CONCLUSION:

In this paper, a deep learning based model is introduced to effectively learn the Glow effect from a night-time scene. Based on this model, a multi-path dilated convolutional network is included that jointly detects and removes the scene glow. The Glow regions are first detected by the DeGlow network which provides additional information for glow removal. Hence, to restore the images captured in night illumination, we introduced a recurrent network that progressively removes the Glow effect and embedded a DeHaze network to remove the atmospheric veils. The experimental evaluation of real and imaginary images demonstrate that our model outperforms the state-of-the-art methods in terms of PSNR and SSIM metrics, and also in computation time. As a future direction, we would like to experiment on an unsupervised generative model with a large un-labeled dataset and analyze the performance in terms of various evaluation metrics.

**AUTHOR INFORMATION:**

1. Shiba Kuanar *Member IEEE,* Department of Electrical Engineering, University of Texas Arlington, Email: shiba.kuanar@mavs.uta.edu

2. K.R. Rao, Fellow IEEE, Department of Electrical Engineering, University of Texas Arlington.

3. Dwarikanath Mahapatra *Member IEEE,* Research Staff Member, IBM Research, Melbourne, Australia

4. Monalisa Bilas, Information Systems, University of Texas at Dallas